\documentclass[final]{cvpr}

\usepackage{times}
\usepackage{epsfig}
\usepackage{graphicx}
\usepackage{amsmath}
\usepackage{amssymb}
\usepackage{bbold}

\usepackage{booktabs} 
\usepackage{multirow}
\usepackage{color, xcolor}
\usepackage{tabularx,verbatim}
\usepackage{xspace}
\usepackage{algorithm}
\usepackage{algorithmic}
\usepackage{setspace}
\usepackage{comment}
\usepackage{bbm}
\usepackage{enumitem}

\usepackage[pagebackref=true,breaklinks=true,colorlinks,bookmarks=false]{hyperref}

\colorlet{dark-blue}{blue!70!black}
\colorlet{dark-green}{green!80!black}
\colorlet{dark-red}{red!80!black}
\hypersetup{
    colorlinks=true,%
    citecolor=dark-blue,%
    filecolor=dark-blue,%
    linkcolor=red,%
    urlcolor=magenta
}

\def\cvprPaperID{868} % *** Enter the CVPR Paper ID here
\def\confYear{CVPR 2021}
\definecolor{mypink}{RGB}{219, 48, 122}

\begin{document}

%%%%%%%%% TITLE
\title{Neighborhood Contrastive Learning for Novel Class Discovery}

\author{Zhun Zhong$^{\textcolor{mypink}{1}}\thanks{Equal contribution}$~, Enrico Fini$^{\textcolor{mypink}{1}*}$, Subhankar Roy$^{\textcolor{mypink}{1,3}}$, Zhiming Luo$^{\textcolor{mypink}{2}}\thanks{Corresponding author}$~, Elisa Ricci$^{\textcolor{mypink}{1,3}}$, Nicu Sebe$^{\textcolor{mypink}{1}}$ \\
 \normalsize{$^{\textcolor{mypink}{1}}$University of Trento~~ $^{\textcolor{mypink}{2}}$Xiamen University~~$^{\textcolor{mypink}{3}}$Fondazione Bruno Kessler}\\
}

\maketitle
% Pages are numbered in submission mode, and unnumbered in camera-ready
% \thispagestyle{empty}

%%%%%%%%% ABSTRACT
\begin{abstract}
In this paper, we address Novel Class Discovery (NCD), the task of unveiling new classes in a set of unlabeled samples given a labeled dataset with known classes. We exploit the peculiarities of NCD to build a new framework, named Neighborhood Contrastive Learning (NCL), to learn discriminative representations that are important to clustering performance. Our contribution is twofold. First, we find that a feature extractor trained on the labeled set generates representations in which a generic query sample and its neighbors are likely to share the same class. We exploit this observation to retrieve and aggregate pseudo-positive pairs with contrastive learning, thus encouraging the model to learn more discriminative representations. 
Second, we notice that most of the instances are easily discriminated by the network, contributing less to the contrastive loss. To overcome this issue, we propose to generate hard negatives by mixing labeled and unlabeled samples in the feature space. We experimentally demonstrate that these two ingredients significantly contribute to clustering performance and lead our model to outperform state-of-the-art methods by a large margin (\eg, clustering accuracy +13\% on CIFAR-100 and +8\% on ImageNet).
\end{abstract}

%%%%%%%%% BODY TEXT
\section{Introduction}

%-----------Illustration of novel class discovery-------------
\begin{figure}[!t]
\centering
\includegraphics[width=0.95\linewidth]{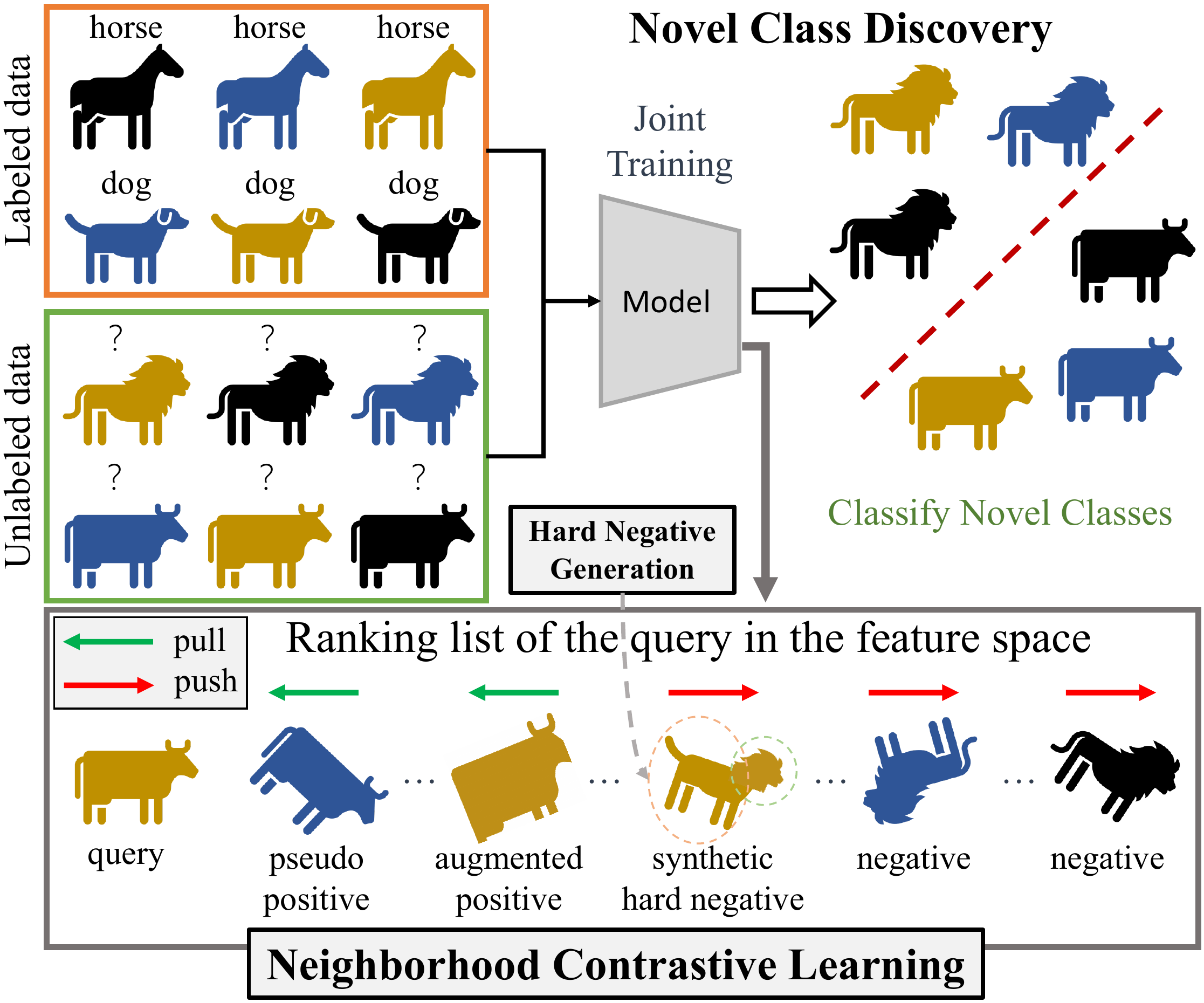}
% \vspace{-.1in}
\caption{Illustration of novel class discovery (NCD) and the proposed neighborhood contrastive learning (NCL). In NCD, we are given two datasets, a labeled one and an unlabeled one, with disjoint class sets.  NCD aims to leverage all data to learn a model that can cluster the unlabeled data. NCL tries to learn discriminative representations by enforcing a query to be close to its correlated view (augmented-positive) and its pseudo-positives (neighbors), as well as to be far from the negatives. We also generate hard negatives by mixing between labeled and unlabeled features, which can further facilitate our NCL.}
% We refer to the classes in the labeled / unlabeled data as old / new classes.
\label{fig:intro-ncd}
% \vspace{-.1in}
\end{figure}
%-------------------

Learning from labeled data has been a widely studied topic in the field of machine learning, and more recently in deep learning~\cite{resnet,krizhevsky2012alexnet,vgg}.
Despite tremendous success, supervised learning techniques largely rely on the availability of massive amounts of annotated data~\cite{deng2009imagenet}. 
To get rid of the difficulty and expensive cost of annotating, 
the machine learning community has shifted the attention to techniques that can learn with limited or completely non-annotated data. To this end, many semi-supervised~\cite{berthelot2019mixmatch,zhang2018mixup} and unsupervised learning~\cite{bautista2016cliquecnn,chen2020simple,he2020moco,xie2016DEC} methods have been proposed, which achieve promising results compared to supervised learning methods. Nonetheless, not much effort has been made to exploit prior knowledge from existing labeled datasets and use it to discover unknown classes that are not present in the labeled data.

In this paper, we address one such relevant problem, called Novel Class Discovery (NCD)~\cite{han2020automatically,han2019learning}, where we are given a labeled dataset and an unlabeled dataset, differing in class label space. 
The goal of NCD is to learn a model that can cluster the unlabeled data by exploiting the latent commonalities from the labeled data (see the top half of Fig.~\ref{fig:intro-ncd}). Importantly, the availability of labeled data does not guarantee transferability because the patterns learned from the labeled data with \textit{off-the-shelf} models might not be useful for the unlabeled data. This poses NCD apart from semi-supervised learning paradigm, where the label space is shared between labeled and unlabeled data, and also makes it more challenging. The NCD task finds relevance in many real-world scenarios where the volume of unlabeled data keeps growing (\textit{e.g.}, multimedia). It is desirable to leverage the existing annotated data (collected from known classes) to explore the new unlabeled data from novel classes, rather than in a completely unsupervised fashion from scratch.

With that goal in mind, this work proposes a holistic learning framework that uses contrastive loss~\cite{he2020moco,tian2019contrastive} formulation to learn discriminative features from both the labeled and unlabelled data, which is absent in most NCD methods~\cite{han2020automatically,han2019learning,hsu2017learning,hsu2019multi}.
Subsequently, we introduce two key ideas in the paper. The first idea is to exploit the fact that the local neighborhood of a sample (\textit{query}) in the embedding space will contain samples which most likely belong to the same semantic category of the query, and can be considered as \textit{pseudo-positives}. Note that this is specific to the NCD setting, where we can pre-train a feature extractor with supervision. We exploit this observation in the context of contrastive learning to bring the query closer to its pseudo-positives, which is termed as \textbf{N}eighborhood \textbf{C}ontrastive \textbf{L}earning (NCL) (see the bottom half of Fig.~\ref{fig:intro-ncd}). These numerous positives allow us to obtain a much stronger learning signal when compared to the traditional contrastive formulation realized with only two views~\cite{chen2020simple, he2020moco}. Our second idea is to address the better selection of \textit{negatives} to further improve the contrastive learning. Peculiar to the NCD task where we are given labeled samples of the known classes (a.k.a \textit{true} negatives of any unlabeled instance), we exploit them, together with the unlabeled samples, to generate synthetic samples in the feature space using a mixing strategy and treat them as \textit{hard} negatives (see Fig.~\ref{fig:intro-ncd}). This circumvents the problem of falsely treating the true positives as negatives~\cite{he2020moco,kalantidis2020hard}. We call this process as \textbf{H}ard \textbf{N}egative \textbf{G}eneration (HNG), which is effective and can produce a boost in performance when employed together with NCL.

To summarize, our contributions are threefold:
\begin{itemize}
    \item We propose Neighborhood Contrastive Learning (NCL) for NCD, which exploits the local neighborhood in the embedding space of a given query. Our NCL recruits more positive samples for the contrastive loss formulation, significantly improving the clustering accuracy.
    \item We propose to aid the contrastive learning by leveraging the labeled samples to generate hard negative samples through feature mixing. With labeled data from various classes, the proposed Hard Negative Generation (HNG) can obtain consistent improvement.
    \item Extensive experiments on three NCD benchmarks demonstrate the effectiveness of our method and show that we advance the state-of-the-art approaches by large margins (\eg, clustering accuracy +13\% on CIFAR-100 and +8\% on ImageNet).
\end{itemize}

%-------------------------------------------------------------------------
% Related works
\section{Related Work}
\label{sec:related}

\textbf{Novel Class Discovery} is a relatively new task that aims to classify the samples in the unlabeled set into different semantic categories. It is different from unsupervised clustering in that one has a labeled set which has completely different classes from the unlabeled set. Typical novel class discovery methods first train a model on the labeled data and use it as an initialization for performing unsupervised clustering on the unlabeled data.
The works~\cite{hsu2017learning, hsu2019multi} in this category utilize the labeled data to train a binary classification model by exploiting the pair-wise similarity of images and then use this trained binary classification model as a supervision for clustering on the unlabeled data. Similarly, \cite{han2019learning} pretrains the model on the labeled data, followed by an end-to-end clustering technique~\cite{xie2016DEC} on the unlabeled data. Deviating from this two-stage training strategy, Han \textit{et. al.}~\cite{han2020automatically} propose to leverage labeled data while performing unsupervised clustering on the unlabeled data. Our proposed NCL also builds on the premise of leveraging labeled data in the unsupervised clustering phase. However, in contrast to~\cite{han2020automatically}, NCL uses labeled data not to maintain the basic discrimination of representation, but to aid the contrastive learning process by generating more informative negatives.

\textbf{Unsupervised Clustering} is the task to partition an unlabeled dataset into different semantic categories, where the prior knowledge of a labeled set is not available. To this end, many shallow~\cite{arthur2006k,macqueen1967_kmeans,zelnik2005self} and deep learning based methods~\cite{Chang_2017_ICCV,ghasedi2017deep,rebuffi2020lsd,van2020scan,xie2016DEC,yang2017towards,zhuang2019local} have been proposed. The deep learning based methods can be roughly categorized into two kinds where the first kind exploits pairwise similarity of the samples to generate pseudo-labels for clustering~\cite{Chang_2017_ICCV,han2020automatically,rebuffi2020lsd}. Whereas, the second kind~\cite{van2020scan,zhuang2019local} uses \textit{neighborhood aggregation} of feature embedding to bring closer the similar instances and simultaneously pushing away the dissimilar instances, thereby achieving a clustering effect. Our method also draws inspiration from these two lines of works. Of notable interest to our work, \cite{zhuang2019local} uses a non-deterministic \textit{k-means} algorithm to find a local neighborhood within an iterative optimization process, which however is sensitive to initialization and also computationally expensive.
Instead, this paper proposes to adopt an end-to-end clustering technique via the use of pairwise similarity of samples and directly explore neighborhood by $k$-nearest neighbors, which makes our method much simpler while still retaining the benefits of neighborhood aggregation.

%-----------framework-------------
\begin{figure*}[!ht]
\centering
\includegraphics[width=0.9\linewidth]{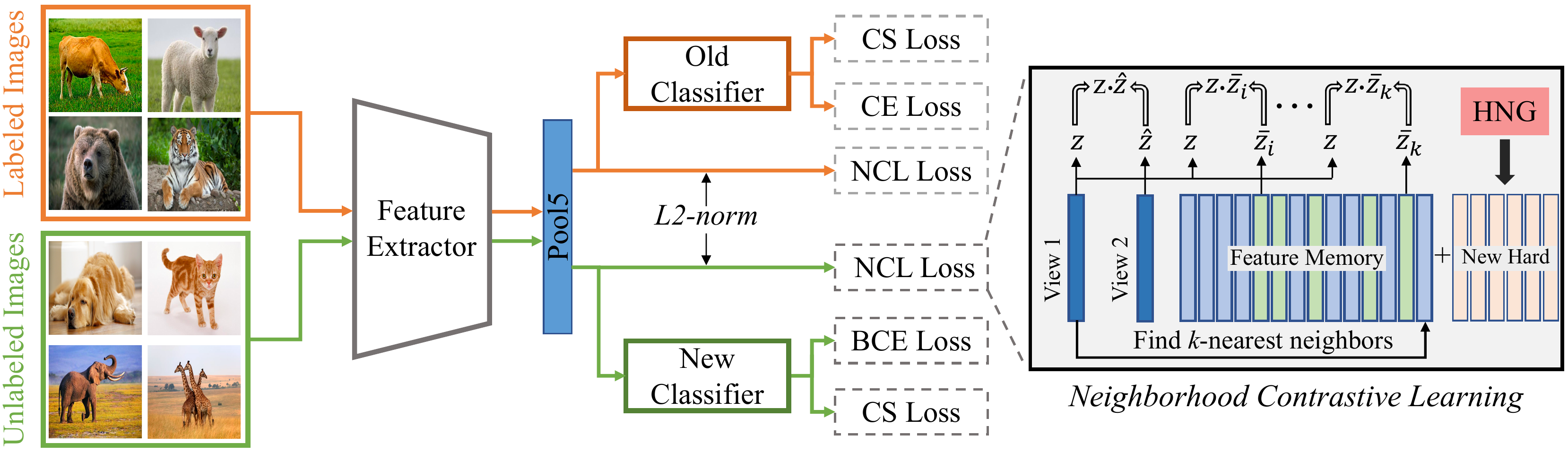}
\vspace{-.05in}
\caption{The proposed neighborhood contrastive learning framework for novel class discovery. Given training images sampled from the labeled and the unlabeled data, we forward them into the network to obtain corresponding representations. For the labeled data, the CE loss, CS loss and the proposed NCL loss are calculated with the ground-truth labels. For the unlabeled data, BCE loss and CS loss are computed to optimize the new classifier while the NCL loss is proposed to learn discriminative representation. \textbf{CE:} cross-entropy, \textbf{BCE:} binary cross-entropy, \textbf{CS:} consistency, \textbf{NCL:} neighborhood contrastive learning, \textbf{HNG:} hard negative generation.}
\label{fig:framework}
\vspace{-.1in}
\end{figure*}
%-------------------

\textbf{Contrastive Learning} is an unsupervised feature representation learning technique that has gained significant momentum in the recent years. The crux of contrastive learning based methods~\cite{bachman2019learning, chen2020simple,hadsell2006dimensionality, he2020moco,tian2019contrastive,wu2018unsupervised} lies in computing a similarity between an input and its correlated view, instead of a fixed target (\textit{e.g.}, one-hot label). Due to the close association between unsupervised learning and NCD, we adopt the contrastive loss~\cite{hadsell2006dimensionality} formulation to harness its power for learning discriminative representations. However, different from the above methods, the contrastive loss formulation in NCL exploits both the labeled data and the unlabeled data into one holistic framework, which is well suited for the NCD task. Moreover, in NCL we propose to amalgamate contrastive learning with neighborhood aggregation by considering $k$-nearest neighbors as pseudo-positives, making our formulation unique in the NCD literature.

\textbf{Negative Mining} plays a crucial role in contrastive learning because the success of the contrastive loss is pivoted on the presence of useful \textit{negatives}~\cite{he2020moco}. Aside from maintaining a large batch size~\cite{chen2020simple} or a large queue~\cite{he2020moco} for having ample useful negatives, one can draw inspirations from the semi-supervised learning literature and naturally consider using mixup strategy in the pixel space~\cite{zhang2018mixup} or the latent space~\cite{verma2019manifold} to generate harder negatives~\cite{kalantidis2020hard,robinson2020contrastive}. We, therefore, capitalize on the fact that the samples of the known classes in the labeled set are \textit{true negatives} (being disjoint to the novel classes) and their mixing with the farthest features in a queue produces \textit{synthetic} features which are considerably true negatives and harder than the farthest features for the query. Importantly, in NCD, due to the large population of positives in the queue, mixing of two random samples may lead to the generation of false negatives, which can indeed hurt the performance. Hence, our hard negative generation strategy (see Sec.~\ref{sec:hng}) alleviates the drawbacks of~\cite{kalantidis2020hard,robinson2020contrastive} and is tailor-made to NCD.
%-------------------------------------------------------------------------

%-------------------------------------------------------------------------
% Method
\section{Method}
\label{sec:method}
\textbf{Problem Definition.}
The task of Novel Class Discovery (NCD) assumes the availability of two datasets: a labeled dataset $D^l$ and an unlabeled dataset $D^u$, containing $C^l$ and $C^u$ classes respectively. The sets of classes in the two datasets are disjoint, but some degree of similarity between the two is necessary. The goal of NCD is to cluster the data in $D^u$, leveraging the knowledge from $D^l$. 

\textbf{Overall Framework.} To discover the latent classes in $D^u$, we learn a shared feature extractor $\Omega: x \mapsto z \in R^H$ and two linear classifiers $\phi^l$ and $\phi^u$, with $C^l$ and $C^u$ output neurons respectively, each followed by a softmax layer. At each training step, a batch of images is sampled from both $D^u$ and $D^l$. Using data augmentation we generate two correlated views of the same batch and forward them into the feature extractor. On the one hand, the features of the labeled images are fed to the classifier $\phi^l$, which is optimized with the cross-entropy loss using the labels. On the other hand, using the binary cross-entropy loss, the classifier $\phi^u$ learns to infer the cluster assignments for the unlabeled images. Both classifiers are encouraged to output consistent predictions using the consistency loss. In addition, the representations $z$ are refined by the proposed neighborhood contrastive loss (NCL) on both labeled and unlabeled samples. 
The overall framework is depicted in Fig.~\ref{fig:framework}.

\subsection{Baseline for Novel Class Discovery} 
\label{baseline}

For the baseline, we use a three stage learning pipeline similar to \cite{han2020automatically}. First, we learn a label-agnostic image representation by self-supervision learning~\cite{gidaris2018rotation} using both labeled and unlabeled datasets, which has been shown to be particularly good at extracting low-level features in the first layers of the network~\cite{asano2019critical}.

Subsequently, high-level features are learned using supervision on the labeled dataset. Given a sample and its label $(x, y) \in D^l$, we optimize the network using the \textit{cross-entropy} loss:
\begin{equation}
%\small
\ell_{ce}=-\frac{1}{C^l} \sum_{i=1}^{C^l} y_{i} \log \phi^{l}_{i}\left(\Omega\left(x \right)\right).
\end{equation}

Finally, we simplify the cluster discovery step in \cite{han2020automatically} by using the cosine similarity of the features to estimate pairwise pseudo-labels, instead of ranking statistics. We find this modification can yield similar performance with respect to ranking statistics when applied with our NCL, while being significantly more efficient and easier to implement.
Specifically, given a pair of images $(x^u_i, x^u_j)$ sampled from dataset $D^u$, we extract features $(z^u_i, z^u_j)$ and compute their cosine similarity $\delta\left(z^u_i, z^u_j\right)=z^{u\top}_i z^u_j /\|z^u_i\|\|z^u_j\|$. The pairwise pseudo-label is then assigned as follows:
\begin{equation}
\label{pairwise-pseudo-label}
%\small
\hat{y}_{i,j}=
\mathbbm{1}\left[ \delta\left(z^u_i, z^u_j\right) \geq \lambda \right],
\end{equation}
where $\lambda$ is a threshold that represents the minimum similarity for two samples to be assigned to the same latent class. Then, the pairwise pseudo-label is compared to the inner product of the outputs of the unlabeled head $p_{i,j} = \phi^{u}\left(z_{i}^{u}\right)^{\top} \phi^{u}\left(z_{j}^{u}\right)$. The network is optimized using the \textit{binary cross-entropy} loss:
\begin{equation}
%\small
\ell_{bce}= - \hat{y}_{i,j} \log\left( p_{i,j} \right) - \left(1-\hat{y}_{i,j}\right) \log (1 - p_{i,j}).
\end{equation}

The last building block of our baseline is the consistency loss, which enforces the network produce similar predictions for an image $x_i$ and its correlated view $\hat{x}_i$. This is particularly important for unlabeled examples. 
Nonetheless, we find that consistency helps with both labeled and unlabeled examples. We use \textit{mean squared error}:
\begin{equation}
    %\small
    \begin{aligned}
        \ell_{mse}=\frac{1}{C^l} \sum_{i=1}^{C^l}\left(\phi_{i}^{l}\left(z^{l}\right)-\phi_{i}^{l}\left(\hat{z}^{l}\right)\right)^{2} + \\ \frac{1}{C^u} \sum_{j=1}^{C^u}\left(\phi^{u}_{j}\left(z^{u}\right)-\phi_{j}^{u}\left(\hat{z}^{u}\right)\right)^{2}.
    \end{aligned} 
\end{equation}
The overall loss for our baseline reads as:
\begin{equation}
    %\small
    \ell_{base} = \ell_{ce} + \ell_{bce} + \omega\left(t\right)\ell_{mse},
\end{equation}
where the coefficient $\omega\left(t\right)$ is a ramp-up function as in \cite{han2020automatically}.

\subsection{Neighborhood Contrastive Learning}
\label{sec:ncl}
Given a set of stochastic image transforms, we generate two correlated views $\left(x^u, \hat{x}^u\right)$ of a generic unlabeled sample to be used as a positive pair. Subsequently, we apply the network $\Omega$ to extract $\left(z^u, \hat{z}^u\right)$ from the views. This operation is repeated for all the samples of a batch of length $B$. We also maintain a queue $M^u$ of features stored from past training steps, which initially are regarded as negatives, denoted with $\bar{z}^u$. The contrastive loss for the positive pair can be written as:
\begin{equation}
\label{contrastive}
    %\small
    \ell_{\left(z^u, \hat{z}^u\right)}=-\log \frac{e^{\delta\left(z^u, \hat{z}^u\right) / \tau}}{e^{\delta\left(z^u, \hat{z}^u\right) / \tau} + \sum_{m=1}^{|M^u|} e^{\delta\left(z^u, \bar{z}^u_m\right) / \tau}},
\end{equation}
where $\delta(\cdot,\cdot)$ is the cosine similarity and $\tau$ is a temperature parameter that controls the scale of distribution.

Unfortunately, a well-known drawback of the contrastive loss is that samples belonging to the same class could be treated as negatives, since we have no information about the labels. However, intuitively, the quality of the representations should benefit if the positive and negative pairs correspond to the desired latent classes. One way to mitigate this issue is to use the model itself to generate pseudo-positive pairs of samples, \textit{i.e.}, to consider the \textit{neighbors} of the representation $z^u$ as instances of the same class. The selection of sensible pseudo-positive pairs turns out to be a hard task, especially at the beginning of the training, when the quality of the representations is poor. However, in NCD, we can leverage the labeled dataset $D^l$ to bootstrap the representations, and then use them to infer the relationships between unlabeled examples in $D^u$.

More formally, given a network $\Omega$ pretrained as the first two steps described in Sec.~\ref{baseline}, we can retrieve the top-$k$ most similar features from the queue for a query $z^u$:
\begin{equation}
\label{eq:topk}
    %\small
    \rho_k = \underset{\bar{z}^u_i}{\operatorname{argtop}_k} \left(\left\{ \delta\left(z^u, \bar{z}^u_i\right) | \, \forall i \in \left\{1, \dots, |M^u| \right\}\right\} \right).
\end{equation}
Assuming the examples in $\rho_k$ are false-negatives (\textit{i.e.}, they belong to the same class as $z^u$), we can regard them as pseudo-positives and write their contributions in the contrastive loss as follows:
\begin{equation}
    \label{eq:pp-contrastive}
    %\small
    \ell_{\left(z^u,\rho_k\right)} = - \frac{1}{k}\sum_{\bar{z}^u_i \in \rho_k}{\log \frac{e^{\delta\left(z^u, \bar{z}^u_i\right) / \tau}}{e^{\delta\left(z^u, \hat{z}^u\right) / \tau} + \sum_{m=1}^{|M^u|} e^{\delta\left(z^u, \bar{z}^u_m\right) / \tau}}}.
\end{equation}
Finally we can introduce our  \textit{Neighborhood Contrastive loss} as follows:
\begin{equation}
    \ell_{ncl}=\alpha\ell_{\left(z^u, \hat{z}^u\right)} + \left(1 - \alpha\right)\ell_{\left(z^u,\rho_k\right)},
\end{equation}
where $\alpha$ controls the weight of the two components.

\subsection{Supervised Contrastive Learning}
In the case of the labeled dataset $D^l$, instead of using the network to mine the pseudo-positives, we can directly use the ground-truth labels to retrieve the set of positives from the queue of labeled set $M^l$ for a sample $x^l_i$ with features $z^l_i$:
\begin{equation}
%\small
\rho = \left\{\bar{z}^l_j \in M^l: y_i=y_j\right\} \cup \hat{z}_i^l.
\end{equation}
Note that $\rho$ contains both the features $\hat{z}_i^l$ of the correlated view $\hat{x}^l_i$ and the other samples belonging to the same class. Using this supervision, our Neighborhood Contrastive loss can be reduced to the \textit{Supervised Contrastive loss}~\cite{khosla2020supervised}:
\begin{equation}
%\small
\ell_{scl}=-\frac{1}{|\rho|} \sum_{\mathring{z}^l_j \in \rho} \log \frac{e^{\delta\left(z^l_i, \mathring{z}^l_j\right) / \tau}}{e^{\delta\left(z^l_i, \hat{z}_i^l\right) / \tau} + \sum_{m=1}^{|M^l|} e^{\delta\left(z^l_i, \bar{z}^l_m\right) / \tau}}.
\end{equation}

\subsection{Hard Negative Generation}
\label{sec:hng}
He et al.~\cite{he2020moco} show the importance of having a large memory that covers a rich set of negative samples for contrastive learning. Recently, other studies \cite{kalantidis2020hard, robinson2020contrastive} find that most of the negatives have very low similarities with the query sample. We experimentally verify that this behavior is also present when contrastive learning is used in the context of Novel Class Discovery (NCD). Specifically, as detailed in Sec.~\ref{eval}, we demonstrate that removing the easiest negatives from the queue does not impact performance, indicating that such negative samples contribute less during training. This is not desirable, because we are wasting memory and computation. On the other hand, selecting hard negatives automatically can be difficult since we have no information about the latent classes in the unlabeled set, and therefore we could end up selecting positive samples. However, in NCD we assume that the set of classes in the labeled and unlabeled sets are disjoint. This entails that all the samples from one set are negatives for the samples of the other set, and vice versa. Inspired by the advancements in regularization techniques using image / feature mixtures \cite{zhang2018mixup, verma2019manifold}, we use this notion to generate hard negatives by mixing labeled and unlabeled samples.

Given a view $x^u$ of an image belonging to the unlabeled set and its representation in the feature space $z^u$, we can select easy negatives by looking at the features with minimal similarity in the queue $M^u$:
\begin{equation}
\label{eq:neg-topk}
    \varepsilon_k = \underset{\bar{z}^u_i}{\operatorname{argtop}_k} \left(\left\{ -\delta\left(z^u, \bar{z}^u_i\right) | \, \forall i \in \left\{1, \dots, |M^u| \right\}\right\} \right).
\end{equation}
Note the negative sign of the similarity. Since the network can confidently distinguish these samples from the query, we can safely assume that they are very likely to be true negatives, \textit{i.e.} they do not belong to the same class as the query. Note that this is in contrast with the recent literature on hard negative mining \cite{kalantidis2020hard, robinson2020contrastive}, which samples hard negatives, incurring in the problem of false-negatives.

\begin{figure}[!t]
\centering
\includegraphics[width=0.98\linewidth]{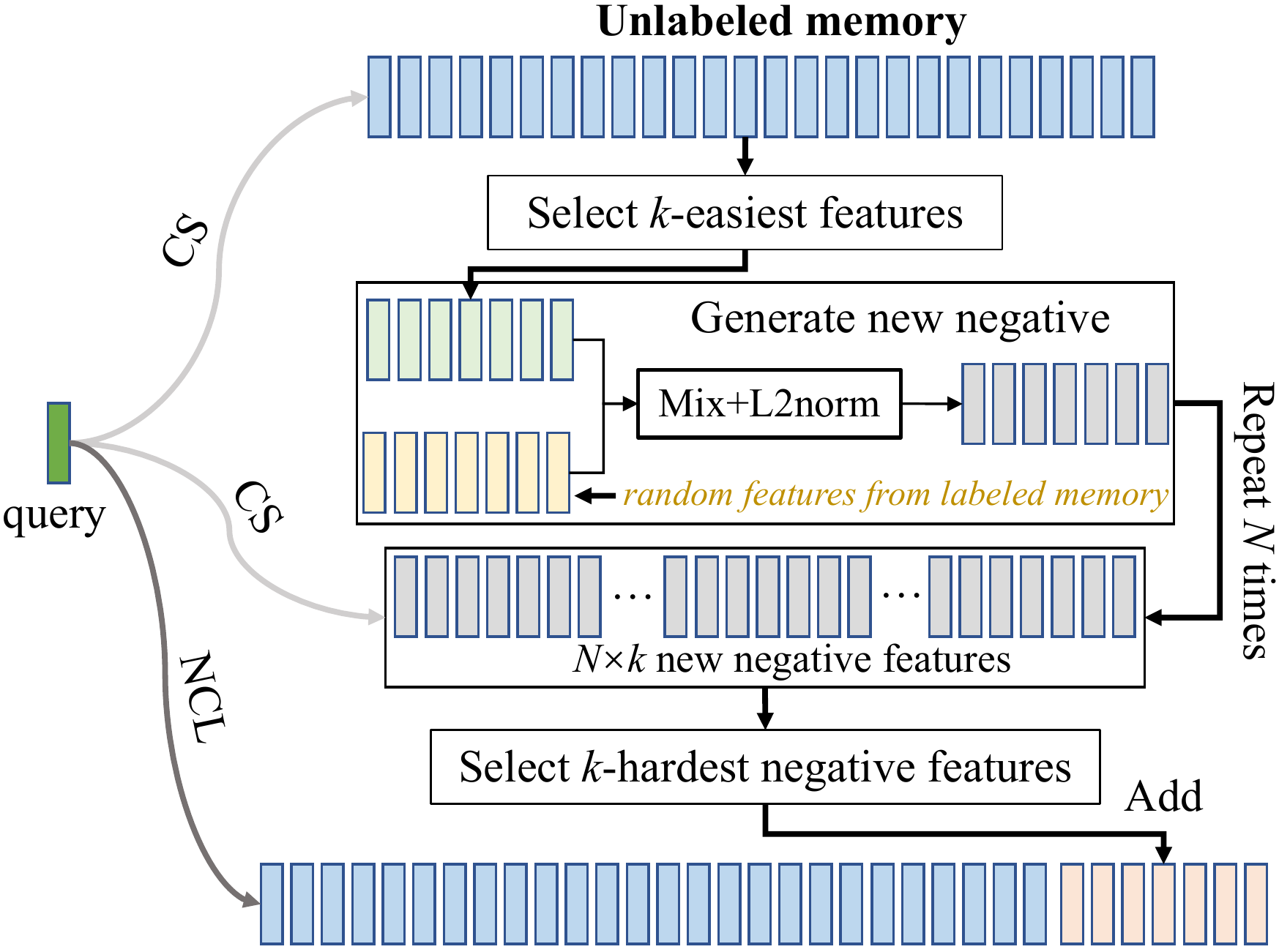}
\vspace{-.03in}
\caption{Illustration of hard negative generation (HNG). \textbf{CS}: compute similarity, \textbf{NCL}: neighborhood contrastive learning.}
\label{fig:illustration-HNG}
\vspace{-.1in}
\end{figure}

Let us also consider a queue $M^l$ containing labeled samples stored from past training steps. As mentioned above, these are by definition true negatives with respect to $x^u$. Our insight is that by linearly interpolating the examples in these two sets we can generate new, hopefully more informative negatives. In practice, for each $\bar{z}^u \in \varepsilon_k$ we randomly sample a feature $\bar{z}^l \in M^l$ and compute the following:
\begin{equation}
\zeta = \mu \cdot \bar{z}^u+(1-\mu) \cdot \bar{z}^l,
\end{equation}
where $\mu$ is the mixing coefficient.
This process of cycling through $\varepsilon_k$ is repeated $N$ times such that the resulting set of mixed negatives $\eta$ will contain $k \times N$ features. Then, the hardest negatives are filtered from $\eta$, using the cosine similarity as before:
\begin{equation}
    \eta_k = \underset{\zeta_i}{\operatorname{argtop}_k} \left(\left\{\delta\left(z^u,\zeta_i\right) | \, \forall i \in \left\{1, \dots, k \times N \right\}\right\} \right).
\end{equation}
This results in a set $\eta_k$ of hard negatives that have the following two properties: (i) they are most likely true negatives, (ii) it is hard for the network to distinguish them from the query.
Finally the queue for $x^u$ is derived by adding the newly generated mixed negatives into the queue $M^u$:
\begin{equation}
    \vspace{-.03in}
    M^{u'} = M^u \cup \eta_k,
    \vspace{-.03in}
\end{equation}
and the contrastive loss is computed as in Eq.~\ref{contrastive} and Eq.~\ref{eq:pp-contrastive}, but replacing $M^u$ with $M^{u'}$. Note that $M^u$ is not overwritten in the process. This pipeline for hard negative generation (illustrated in Fig.~\ref{fig:illustration-HNG}) is repeated for each unlabeled sample in the current batch, as they will have different sets of easy negatives. To distinguish between the number of pseudo-positives used in Eq.~\ref{eq:topk} and number of negatives used in Eq.~\ref{eq:neg-topk}, we denote the former as $k_1$ and the latter as $k_2$ respectively.

%-----------hard negative generation-------------
%-------------------

\subsection{Overall Loss}
Considering the baseline model, neighborhood contrastive learning on unlabeled data, supervised contrastive learning on labeled data, and the hard negative generation on unlabeled data, the overall loss for our model is:
\begin{equation}
    \ell_{all} = \ell_{base} + \ell_{ncl} + \ell_{scl}.
\end{equation}
Throughout the paper, we refer to the $\ell_{ncl}$ and $\ell_{scl}$ collectively as neighborhood contrastive learning.

%-------------------------------------------------------------------------
%-------------------------------------------------------------------------

%-------------------------------------------------------------------------
% Experiment

\section{Experiments}
\label{sec:exp}
\subsection{Dataset and Experimental Details}

\textbf{Dataset}. We conduct experiments on three datasets that are commonly used in NCD: CIFAR-10~\cite{krizhevsky2009learning}, CIFAR-100~\cite{krizhevsky2009learning} and ImageNet~\cite{deng2009imagenet}. Following \cite{han2020automatically}, we split the training data of each dataset into a labeled set and an unlabeled set, and assume that the the number of classes in the unlabeled set is known. The partitions of the three datasets are reported in Table~\ref{tabel:data-details}. More details on the datasets can be found in supplementary. Following \cite{han2020automatically,han2019learning}, we report results averaged over 10 runs for CIFAR-10 and CIFAR-100. For ImageNet, we report results averaged over 3 runs using three different unlabeled subsets.

%------------------------------------------------------------------------
\begin{table}[!t]
  \footnotesize
  \centering
  \newcolumntype{C}{>{\centering\arraybackslash}X}%
  \newcolumntype{R}{>{\raggedleft\arraybackslash}X}%
  \begin{tabularx}{0.8\linewidth}{l|CC|Cc}
    \hline
    \multirow{2}{*}{Dataset}   &  \multicolumn{2}{c|}{Labeled Set}   & \multicolumn{2}{c}{Unlabeled Set} \\
               &  \#image & \#class   & \#image & \#class         \\
    \hline
    CIFAR-10 & 25K & 5 & 25K & 5\\
    CIFAR-100 & 40K & 80 & 10K & 20\\
    ImageNet & 1.25M & 882 & $\approx$30K & 30\\
    \hline
  \end{tabularx}
  \caption{Dataset statistics for novel class discovery.}
  \label{tabel:data-details}
  \vspace{-.1in}
\end{table}
%------------------------------------------------------------------------

\textbf{Evaluation Metric}. We employ average clustering accuracy (ACC) to evaluate the performance of different methods on unlabelled data. The ACC is defined as:
\begin{equation}
\vspace{-.05in}
\operatorname{ACC} = \max _{perm \in P}
\frac{1}{N} \sum_{i=1}^{N}
\mathbbm{1}\left \{
  {y}_{i}=perm\left(\hat{y}_{i}\right)
\right \},
\vspace{-.05in}
\end{equation}
where $y_i$ and $\hat{y}_{i}$ represent the ground-truth label and clustering predicted label of a sample $x^u_{i} \in D^u$, respectively. $P$ is the set of all permutations, which can be rapidly computed by the Hungarian algorithm~\cite{kuhn1955hungarian}.

\textbf{Implementation Details}. For a fair comparison with existing methods, we use ResNet-18 \cite{resnet} as the backbone of our method for all three datasets. We follow \cite{han2020automatically} to initialize the model with self-supervised learning on the whole data and fine-tune the model with supervised learning on the labeled data, more training details can be found in \cite{han2020automatically}. In the step of novel class discovery on the unlabeled data, we use SGD optimizer to update the network. Note that, in the steps of supervised fine-tuning and novel class discovery, we only update the last convolutional block of the ResNet and the two classifiers. The initial learning rate is set to 0.1 and is divided by 10 after every 170/30 epochs for \{CIFAR-10, CIFAR-100\}/ImageNet. We train the model with 200/90 epochs in total for \{CIFAR-10, CIFAR-100\}/ImageNet. We randomly sample training samples from both the labeled and unlabeled data, where the batch size is set to 128/512 for \{CIFAR-10, CIFAR-100\}/ImageNet. For the consistency loss, we apply the ramp-up function with weight $\gamma=\{5, 50, 10\}$ and ramp-up length $T=\{50, 150, 50\}$ for CIFAR-10, CIFAR-100 and ImageNet, respectively. For the binary-cross entropy loss, we set $\lambda=0.95$. %

For our method, we introduce the neighborhood contrastive learning (NCL) and hard negative generation (HNG) at the 2\textit{th} and 4\textit{th} epoch, respectively. In default, we set memory size $|M|$ = 2,000, temperature $\tau$ = 0.05, number of pseudo-positives $k_1$ = $|M|/C^u/2$, weight of augmented-positive $\alpha$ = 0.2, number of negative samples $k_2$ = 400, and number of HNG iterations $N$ = 5. For each mixing process, we generate new negatives with $\mu$ = 1/3 and $\mu$ = 2/3. That is, each mixing process will be performed twice using these two values of $\mu$. We find the above parameter settings can consistently achieve stable and well performance across datasets. The parameter analysis can be found in the supplementary material.

\subsection{Evaluation} \label{eval}

%------------------------------------------------------------------------
\begin{table}[!t]
  \small
  \centering
  \newcolumntype{C}{>{\centering\arraybackslash}X}%
  \newcolumntype{R}{>{\raggedleft\arraybackslash}X}%
  \begin{tabularx}{0.9\linewidth}{l|C|C}
    \hline
    Method   &   CIFAR-10   & CIFAR-100 \\
    \hline
    Basel. w/o SSL & 85.0$\pm$0.4\%& 66.5$\pm$4.0\%\\
    Basel. w/o CE & 83.9$\pm$9.4\%& 62.6$\pm$3.6\% \\
    Basel. w/o BCE & 39.5$\pm$4.2\%& 18.1$\pm$0.8\%\\
    Basel. w/o CS & 84.1$\pm$0.9\%& 61.6$\pm$3.2\%\\
    \bf Baseline & \bf 87.9$\pm$0.7\%& \bf 69.4$\pm$1.4\%\\
    \hline
  \end{tabularx}
  \caption{Ablation study of the baseline method on CIFAR-10 and CIFAR-100. \textbf{SSL}: self-supervised learning, \textbf{CE}: cross-entropy loss on the labeled data, \textbf{BCE}: binary cross-entropy loss on the unlabeled data, \textbf{CS}: consistency loss.}
  \label{tabel:ablation-on-baseline}
\end{table}
%------------------------------------------------------------------------

\textbf{Ablation study on the baseline}. We first evaluate the effectiveness of the components of the baseline, including self-supervised learning (SSL), cross-entropy (CE) loss on the labeled data, binary cross-entropy (BCE) loss on the unlabeled data, and consistency (CS) loss. We individually remove each of them from the baseline and evaluate the performance. Results are reported in Table~\ref{tabel:ablation-on-baseline}. We mainly make the following four observations: (1) Removing each component will reduce the results of the baseline. (2) BCE is the most important component. When removing BCE, the results decrease substantially. Without BCE, the classifier is only learned with a weak supervision (\textit{i.e.}, consistency loss) and therefore fails to cluster the samples. (3) Removing SSL from the baseline will  decrease the performance. This is due to the fact that SSL improves the generality of the representations and thus benefits the learning of the BCE. (4) CS is also beneficial in novel class discovery, since it encourages the classifier to be more robust to intra-class variations. The above observations verify the effectiveness and importance of each component in the baseline.

%------------------------------------------------------------------------
\begin{table}[!t]
  \footnotesize
  \centering
  \newcolumntype{C}{>{\centering\arraybackslash}X}%
  \newcolumntype{R}{>{\raggedleft\arraybackslash}X}%
  \begin{tabularx}{0.95\linewidth}{l|c|c}
    \hline
    Method   &   CIFAR-10   & CIFAR-100 \\
    \hline
    Baseline & 87.9$\pm$0.7\%& 69.4$\pm$1.4\%\\
    \hline
    + NCL w/o PP & 61.8$\pm$7.6\%~\textcolor{dark-red}{($\downarrow$ 26.1\%)}& 68.5$\pm$1.9\%~\textcolor{dark-red}{($\downarrow$ 0.9\%)}\\
    + NCL w/o AP & 90.9$\pm$2.1\%~\textcolor{dark-green}{($\uparrow$ 3.0\%)}&79.7$\pm$5.7\%~\textcolor{dark-green}{($\uparrow$ 10.3\%)}\\
    + NCL w/o LA &
    93.3$\pm$0.1\%~\textcolor{dark-green}{($\uparrow$ 5.4\%)}&80.3$\pm$0.9\%~\textcolor{dark-green}{($\uparrow$ 10.9\%)}\\
    + NCL & 93.4$\pm$0.2\%~\textcolor{dark-green}{($\uparrow$ 5.5\%)}& 82.3$\pm$2.6\%~\textcolor{dark-green}{($\uparrow$ 12.9\%)}\\
    \hline
    + NCL + HNG & 93.4$\pm$0.1\%~\textcolor{dark-green}{($\uparrow$ 5.5\%)}& 86.6$\pm$0.4\%~\textcolor{dark-green}{($\uparrow$ 17.2\%)}\\
    \hline
  \end{tabularx}
  \caption{Evaluation of the effectiveness of the proposed neighborhood contrastive learning (NCL) and hard negative generation (HNG). \textbf{NCL w/o PP}: NCL without pseudo-positives, \textbf{NCL w/o LA}: without applying NCL on labeled data. \textbf{NCL w/o AP}: removing augmented-positive during NCL.}
  \label{tabel:ablation-on-ours}
\end{table}
%------------------------------------------------------------------------

\textbf{Evaluation of the neighborhood contrastive learning}. To study the effectiveness of neighborhood contrastive learning (NCL), we implement NCL in four ways. 1) NCL: The proposed NCL. 2) NCL w/o PP: NCL without pseudo-positives, which reduces to the vanilla contrastive learning; 3) NCL w/o LA: NCL without enforcing contrastive learning on the labeled data; 4) NCL w/o AP: NCL without approaching a query to its augmented-positive. Results on CIFAR-10 and CIFAR-100 are reported in Table~\ref{tabel:ablation-on-ours}. First, without neighborhood mining, the model will regard all the positive features in the memory as negative samples and push the query sample far away from its positive features, which will certainly damage the performance. Second, implementing NCL on the labeled data can help improve the discrimination of the model, which can facilitate the process of neighborhood mining and thus improve the ACC, especially given a larger labeled dataset (CIFAR-100). Third, the augmented-positive sample is important to improve the performance since it can mitigate the influence caused by the negative samples that are included in the selected KNNs. Fourth, our proposed NCL significantly improves the ACC of baseline. Specifically, NCL gains +5.5\% on CIFAR-10 and +12.9\% on CIFAR-100, demonstrating the effectiveness of the proposed NCL.

%- ----------fig: HNG evaluation-------------
\begin{figure}[!t]
\centering
\includegraphics[width=0.85\linewidth]{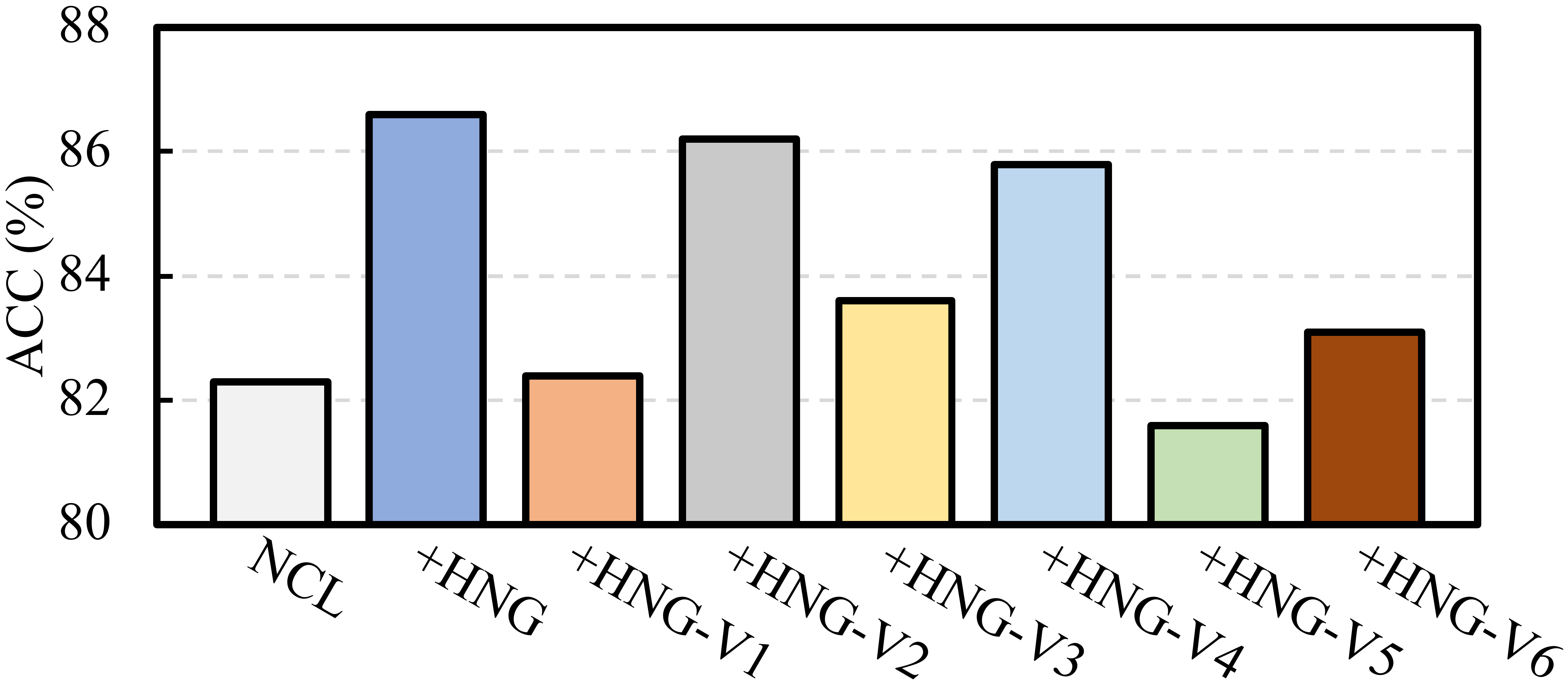}
% \vspace{-.1in}
\caption{Comparison of the proposed hard negative generation (HNG) and its variants on CIFAR-100.}
\label{fig:nhg-evaluation}
% \vspace{-.1in}
\end{figure}
%-------------------

\textbf{Evaluation of hard negative generation}. We first evaluate our proposed hard negative generation (HNG) in Table~\ref{tabel:ablation-on-ours}. We find that HNG significantly increases the ACC for CIFAR-100. However, there is no boost for CIFAR-10. This is likely due to the fact that the labeled set in CIFAR-10 contains a small number of classes. In such a context, mixing between labeled and unlabeled samples is unable to generate diverse hard negative samples and thus fails to facilitate contrastive learning. In Table~\ref{tabel:state-of-the-art}, we show that HNG can also improve the ACC for ImageNet, where the labeled set contains a large amount of classes. This further verifies the effectiveness of our HNG when given a rich labeled dataset. Another beneficial side-effect of HNG is the fact that it helps in stabilizing the training, reducing the variance of the results across all datasets (see Table~\ref{tabel:state-of-the-art}).

To further study the advantage of our HNG, we compare HNG with 6 variants and, based on the results in Fig.~\ref{fig:nhg-evaluation} we make the following observations.
\noindent (\textbf{HNG-V1}): Directly removing $k$-easiest unlabeled samples when computing NCL for each query rarely affects the ACC, supporting our point that easy negative samples contribute less to contrastive learning;
\noindent (\textbf{HNG-V2}): Replacing $k$-easiest unlabeled samples with generated hard negative samples produces similar ACC to directly adding generated hard features to the feature queue (HNG);
\noindent (\textbf{HNG-V3}): Directly using $k$ randomly selected labeled samples as hard negative samples can slightly improve the ACC;
\noindent (\textbf{HNG-V4}): Generating hard negative samples by mixing only on $k$ randomly selected labeled samples can achieve further improvement over ``HNG-V3'';
\noindent (\textbf{HNG-V5}): Generating hard negative samples by mixing only on $k$-easiest unlabeled samples fails to improve the performance;
\noindent (\textbf{HNG-V6}): Generating hard negative samples by mixing on $k$-easiest unlabeled samples and $k$-nearest labeled samples is suboptimal w.r.t using randomly selected labeled feature (HNG). This is because the $k$-nearest labeled features mostly are of the same class, limiting the variety of the generated hard features.

Taking the above observations, the proposed HNG can generate more variety and hard negative samples, which are effective in improving contrastive learning.

\begin{table*}[!t]
  \centering
  \small
  \newcolumntype{C}{>{\centering\arraybackslash}X}%
  \newcolumntype{R}{>{\raggedleft\arraybackslash}X}%
  \begin{tabularx}{0.68\linewidth}{l|c|C|C|c}
    \hline
    Method   &          Venue                            & CIFAR-10        & CIFAR-100       & ImageNet \\
    \hline
    \multicolumn{5}{c}{\textit{Methods without self-supervised learning}} \\
    \hline
    \hline
    $k$-means~\cite{macqueen1967_kmeans} & Classic & 65.5$\pm$0.0\%          & 56.6$\pm$1.6\%          & 71.9\%\\
    KCL~\cite{hsu2017learning}  & ICLR'18              & 66.5$\pm$3.9\%          & 14.3$\pm$1.3\%          & 73.8\% \\
    MCL~\cite{hsu2019multi}     & ICLR'19           & 64.2$\pm$0.1\%          & 21.3$\pm$3.4\%          & 74.4\% \\
    DTC~\cite{han2019learning}  & ICCV'19       & 87.5$\pm$0.3\%          & 56.7$\pm$1.2\%          & 78.3\%\\
    \hline
    \multicolumn{5}{c}{\textit{Methods with self-supervised learning}} \\
    \hline
    \hline
    $k$-means~\cite{macqueen1967_kmeans}$^*$& Classic
         & 72.5$\pm$0.0\%                               & 56.3$\pm$1.7\%          & 71.9\% \\
    KCL~\cite{hsu2017learning}$^*$& ICLR'18
         & 72.3$\pm$0.2\%                               & 42.1$\pm$1.8\%          & 73.8\%\\
    MCL~\cite{hsu2019multi}$^*$ & ICLR'19
         & 70.9$\pm$0.1\%                               & 21.5$\pm$2.3\%          & 74.4\% \\
    DTC~\cite{han2019learning}$^*$ & ICCV'19
         &88.7$\pm$0.3\%  & 67.3$\pm$1.2\%      & 78.3\% \\
    RS~\cite{han2020automatically}$^*$ & ICLR'20 & 90.4$\pm$0.5\% & 73.2$\pm$2.1\% & 82.5\% \\
    \hline
    \bf Ours$^*$ w/o HNG & CVPR21 & \bf 93.4$\pm$0.2\% & \bf 82.3$\pm$2.6\%& \bf 89.5\% \\
    \bf Ours$^*$ & CVPR21 & \bf 93.4$\pm$0.1\% & \bf 86.6$\pm$0.4\%& \bf 90.7\% \\
    \hline
  \end{tabularx}
  \caption{Comparison with state-of-the-art methods on CIFAR-10, CIFAR-100 and ImageNet for novel class discovery. Clustering accuracy is reported on the unlabelled set. ``*'' indicates methods that initialize models with self-supervised learning, except when evaluated on ImageNet. \textbf{Ours}: our method with both neighborhood contrastive learning and hard negative generation, \textbf{Ours w/o HNG}: our method without hard negative generation.}
  \label{tabel:state-of-the-art}
\end{table*}

%-----------visualization-------------
\begin{figure*}[!t]
\centering
\vspace{-.05in}
\includegraphics[width=0.88\linewidth]{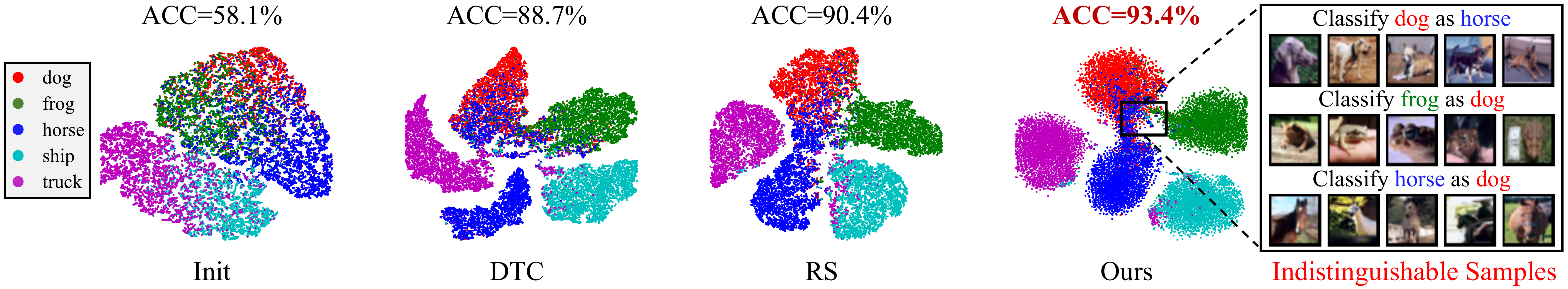}
% \vspace{.05in}
\caption{Feature visualization on CIFAR-10. We extract the output of the last pooling layer as the feature for all unlabeled data and use \textit{t}-SNE~\cite{TSNE} to map the features into a 2D embedding space. We compare our method with the initialized model (trained only with self-supervised learning and supervised learning), DTC~\cite{han2019learning} and RS~\cite{han2020automatically}. We also show examples that are visually similar to samples of other classes and are classified to wrong classes.}
\label{fig:tsne}
\vspace{-.1in}
\end{figure*}
%-------------------

\subsection{Comparison with State-of-The-Art Methods}

We compare the proposed approach with one classical method ($k$-means~\cite{macqueen1967_kmeans}) and four state-of-the-art methods (\textit{i.e.}, KCL~\cite{hsu2017learning}, MCL~\cite{hsu2019multi}, DTC~\cite{han2019learning} and RS~\cite{han2020automatically}). For method based on  $k$-means~\cite{macqueen1967_kmeans}, we first use the labeled data to pre-train the model by supervised learning loss (\textit{i.e.}, cross-entropy loss). Then, we use the trained model to extract features for the unlabeled data without further learning on the unlabeled data. Finally, we perform $k$-means clustering on these extracted features to obtain the clustering results. Except RS~\cite{han2020automatically}, all the other compared methods do not apply self-supervised learning technique. In order to make a fair comparison, we implement these methods (except RS~\cite{han2020automatically}) with two settings depending on whether to utilize self-supervised learning to pre-train the model. With self-supervised learning, we first initialize the model by the rotation loss~\cite{gidaris2018rotation} using both labeled data and unlabeled data and then implement the methods with their own algorithms. Note that, since ImageNet has sufficient training samples from various classes, we directly use the labeled data to pre-train the model with cross-entropy loss for both settings. Comparison results are reported in Table~\ref{tabel:state-of-the-art}.

We can obtain the following two conclusions. First, using self-supervised learning generally can improve the results of all methods, except when evaluated $k$-means~\cite{macqueen1967_kmeans} on CIFAR-100. For example, when using self-supervised learning, the ACC of KCL~\cite{hsu2017learning} is increased from 66.5\% to 72.3\% and from 14.3\% to 42.1\% on CIFAR-10 and CIFAR-100, respectively.
This indicates the effectiveness of self-supervised learning. Second, two versions of our method outperform the state-of-the-art methods (whether using self-supervised learning or not) by a large margin on all datasets, especially on CIFAR-100 and ImageNet. Specifically, our full method achieves \textbf{ACC=93.4\%} on CIFAR-10, \textbf{ACC=86.6\%} on CIFAR-100 and \textbf{ACC=90.7\%} on ImageNet, respectively. These results are higher than the current best method (RS~\cite{han2020automatically}) by +3\% on CIFAR-10, +13.4\% on CIFAR-100 and +8.2\% on ImageNet, respectively. This demonstrates that our method produces the new state-of-the-art results for novel class discovery.

\subsection{Visualization}
% \vspace{-.08in}
To better understand the proposed method, we visualize the feature embeddings of the unlabeled samples on CIFAR-10 using \textit{t}-SNE~\cite{TSNE}. In Fig.~\ref{fig:tsne}, we compare our method with the initial model and two state-of-the-art methods (DTC~\cite{han2019learning} and RS~\cite{han2020automatically}). The initial model is trained with self-supervised learning on all data and supervised learning on the labeled data. As we can see, the initial model can roughly separate samples into 5 clusters. However, there are also many samples of different classes clustered together, resulting in low clustering accuracy (ACC=58.1\%). Compared to the initial model, the other three methods (DTC, RS and our method) generate more discriminative representations, which produce significantly better clustering results. Since DTC, RS and our method all achieve very high clustering results (ACC$>$88\%), we cannot observe obvious difference in clustering visualization between them. However, for our method, the samples of the same class are mostly clustered in a circular area, which is mainly caused by the constraint of enforcing neighbors to be close. We also show some indistinguishable samples that are located at the class decision boundaries. We find that these samples are visually similar, such as in terms of color (frog and dog) and pose (dog and horse), leading the model fail to distinguish them.
%-------------------------------------------------------------------------

\section{Conclusion}
In this paper, we propose a holistic learning framework for Novel Class Discovery (NCD), which adopts contrastive learning to learn discriminate features with both the labeled and unlabeled data. Specifically, we propose the Neighborhood Contrastive Learning (NCL) to effectively leverage the local neighborhood in the embedding space, enabling us to take the knowledge from more positive samples and thus improve the clustering accuracy. In addition, we also introduce the Hard Negative Generation (HNG), which leverages the labeled samples to produce informative hard negative samples and brings further advantage to NCL. Experiments on three datasets demonstrate the significant superiority of our method over state-of-the-art NCD methods.

% \section*{Acknowledgements}
\vspace{.1in}
{\noindent\textbf{Acknowledgements}}  This work is supported by the EU H2020 SPRING No. 871245 and AI4Media No. 951911 projects; the Italy-China collaboration project TALENT:2018YFE0118400; and the Caritro Deep Learning Lab of the ProM Facility of Rovereto. Zhun Zhong thanks Wenjing Li for encouragement.

% change the section to A, B, C
\renewcommand\thesection{\Alph{section}}
\renewcommand\thefigure{\Alph{figure}}
\renewcommand\thetable{\Alph{table}}
\setcounter{section}{0}
\setcounter{figure}{0}
\setcounter{table}{0}

\section*{Appendix}

%-----------fig: ncl parameter-------------
\begin{figure*}[!t]
\centering
\includegraphics[width=0.99\linewidth]{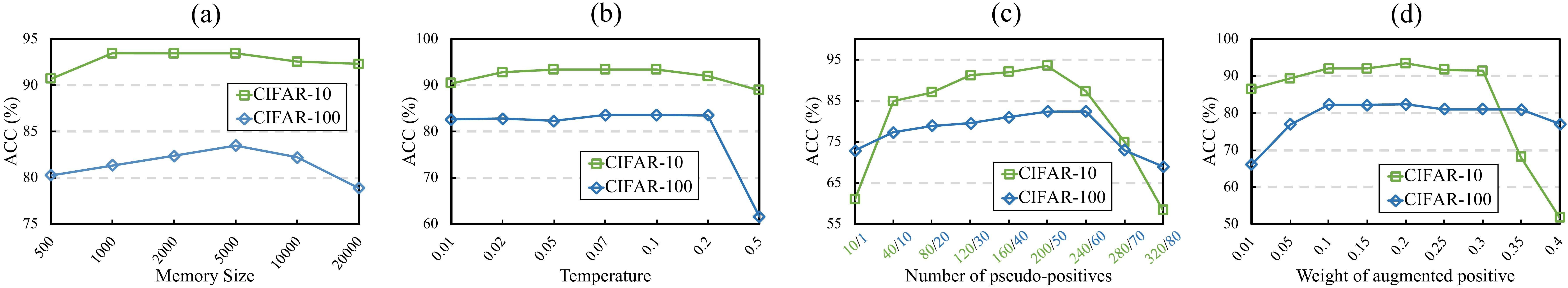}
% \vspace{-.1in}
\caption{Parameter analysis of the proposed neighborhood contrastive learning on CIFAR-10 and CIFAR-100. Sensitivities to (a) memory size, (b) temperature of contrastive learning, (c) number of pseudo-positives, and (d) weight of augmented-positive.}
\label{fig:ncl-parameter}
\vspace{-.1in}
\end{figure*}
%-------------------

\section{Dataset}
\label{sec:dataset}

\textbf{CIFAR-10}~\cite{krizhevsky2009learning} contains 50,000 training images from 10 classes, each of which has a size of $32 \times 32$. For the setting of novel class discovery, we split the samples of the first five classes (namely airplane, automobile, bird, cat and deer) as the labeled data and the remaining samples of the other five classes as the unlabeled data.

\textbf{CIFAR-100}~\cite{krizhevsky2009learning} has the same number of training images and the same image size as CIFAR-10. The difference is that CIFAR-100 is captured from 100 classes. For evaluation, we regard the samples of the first 80 classes as the labeled data and the remaining samples of the other 20 classes as the unlabeled data.

\textbf{ImageNet}~\cite{deng2009imagenet} is a large-scale dataset, including 1.28 million training images from 1,000 classes. Following \cite{hsu2017learning,hsu2019multi}, we divide the training data into two splits that are composed of images from 882 classes and 118 classes, respectively. The split with 882 classes is regarded as the labeled data. For the unlabeled data, we randomly sample three subsets from another split with 118 classes. Each subset consists about 30,000 images from 30 classes and is used as an unlabeled data.

\section{Parameter Analysis}
\label{sec:parameter}
\subsection{Parameter Analysis of NCL}

We investigate four important parameters in neighborhood contrastive learning (NCL), \textit{i.e.}, memory size $|M|$, temperature of contrastive learning $\tau$, number of pseudo-positives $k_1$, and weight of augmented-positive $\alpha$. For evaluation, we vary one parameter at a time while the other three are set to their default values. Results on CIFAR-10 and CIFAR-100 are shown in Fig.~\ref{fig:ncl-parameter}. 

\textbf{(1) Sensitivity to memory size}. In Fig.~\ref{fig:ncl-parameter}(a), we vary the memory size $|M|$ in the range $[500, 20000]$. The ACC first increases with the memory size and achieves the best results when memory size is between 2,000 and 10,000. Considering the efficiency, we set memory size to 2,000, which achieves reasonably good performance on both datasets with limited computational cost (computing the similarities between mini-batch samples and memory samples). 

\textbf{(2) Sensitivity to temperature}. We vary the temperature $\tau$ in the range of [0.01, 0.5] and show the results in Fig.~\ref{fig:ncl-parameter}(b). We can observe that results are similar when temperature is between 0.02 and 0.1, indicating our NCL is robust to temperature within certain ranges. The best results are obtained when temperature is around 0.05. 

\textbf{(3) Sensitivity to number of pseudo-positives}. Since the number of novel classes ($C^u$) is different in CIFAR-10 and CIFAR-100, we vary the number of pseudo-positives $k_1$ in different ranges for these two datasets. The range of $k_1$ is [10, 360] for CIFAR-10 and is [1, 80] for CIFAR-100, respectively. Results are shown in  Fig.~\ref{fig:ncl-parameter}(c). Selecting too few KNNs will ignore most of the positive samples and regard them as negative samples, resulting in worse performance. On the other hand, assigning too many KNNs will include more negative samples. Enforcing a sample to approach too many negative samples will suppress the benefit of true positive samples in the KNNs and will undoubtedly hamper the ACC. Interesting, we find that our NCL achieves consistent good performance when the number of KNNs is equal to the half of $|M|/C^u$ (\textit{i.e.}, 200 for CIFAR-10 and 50 for CIFAR-100).

\textbf{(4) Sensitivity to weight of augmented-positive}. As shown in Fig~.\ref{fig:ncl-parameter}(d), both datasets achieves best results when the weight of augmented-positive $\alpha$ is around 0.2. The ACC will be largely reduced when $\alpha \ge 0.35$ for CIFAR-10. For CIFAR-100, when $\alpha \le 0.05$, the ACC are clearly lower than those of $\alpha \ge 0.1$.

Based on the above analyses, we set the memory size $|M|$=2,000, temperature $\tau$=0.05, number of pseudo-positives $k_1$=$|M|/C^u/2$, and weight of augmented-positive $\alpha$= 0.2 for all datasets in default.

%-----------fig: HNG parameter-------------
\begin{figure}[!t]
\centering
\includegraphics[width=0.99\linewidth]{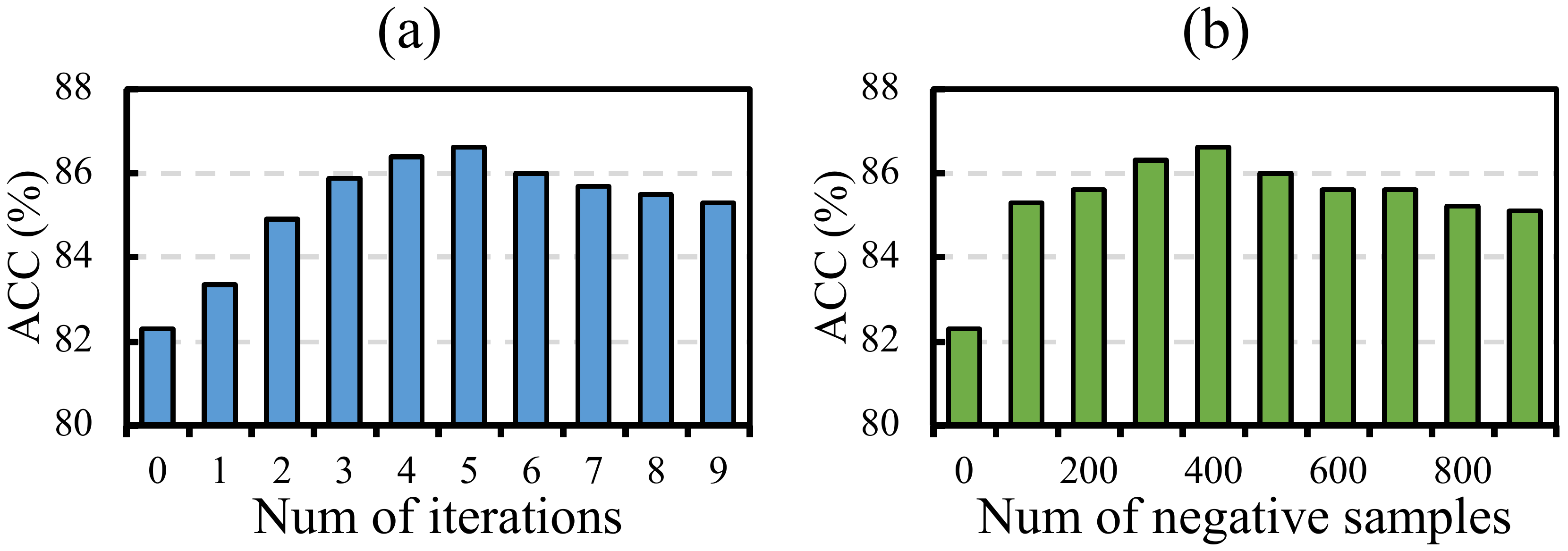}
% \vspace{-.1in}
\caption{Parameter analysis of the proposed hard generative generation (HNG) on CIFAR-100. Sensitivities to (a) number of iterations $N$, and (b) number of negative samples $k_2$.}
\label{fig:nhg-parameter}
% \vspace{-.1in}
\end{figure}
%-------------------

\subsection{Parameter Analysis of HNG}

We evaluate two parameters for hard negative generation (HNG), \textit{i.e.}, number of iterations $N$ and number of negative samples $k_2$. For evaluation, we vary one parameter and fix the other one to its default value. Results on CIFAR-100 are shown in Fig.~\ref{fig:nhg-parameter}. When $N=0$ (in Fig.~\ref{fig:nhg-parameter}(a)) or $k_2=0$ (in Fig.~\ref{fig:nhg-parameter}(b)), the model reduces to the baseline trained only with NCL that does not consider the hard negative samples. As shown in Fig.~\ref{fig:nhg-parameter}, all values of $N$ and $k_2$ achieve higher ACC than the model trained only with NCL, demonstrating the effectiveness of the proposed HNG. The ACC first increases with $N$ /  $k_2$ and achieves best results when $N \approx 5$ /  $k_2 \approx 400$. Performing the HNG with too many iterations or selecting too many negative samples does not lead to further improvement. Considering the above factors, we set $N = 5$ and  $k_2 = 400$ for all datasets in default.

\section{Discussion of Our Method}
\label{sec:dis}

\subsection{Different Impact on CIFAR-10 and CIFAR-100}

From Table~\ref{tabel:dis}, we find that removing pseudo-positives (NCL w/o PP) and adding hard negative generation (NCL + HNG) have inconsistent effects on performance between CIFAR-10 and CIFAR-100. We conjecture that this phenomenon is caused by the difference of number of labeled classes $C^l$ and number of unlabeled classes $C^u$. 1) We have $C^u$=5 for CIFAR-10 and $C^u$=20 for CIFAR-100. NCL w/o PP will regard much more positives as negatives in the memory for CIFAR-10 than CIFAR-100, and thus the performance of CIFAR-10 will be degraded more than of CIFAR-100. 2) We have $C^l$=5 for CIFAR-10 and $C^l$=80 for CIFAR-100. CIFAR-10 contains a small $C^l$. In this context, mixing between labeled and unlabeled samples cannot generate diverse hard negative samples and thus fails to facilitate contrastive learning.

\subsection{Positive Selection for BCE and NCL}

In our method, we use different strategies to select positives for BCE and NCL. The decision is mainly dependent on the number of samples in the batch and memory. 1) The number of samples in batch is much smaller than the size of the memory bank, so the class-balance cannot be ensured. When there are few or no samples of class-$i$ in the batch, we may select overmuch false-positives for a sample of class-$i$ by top-$k$, which will hamper the performance. 2) Using the memory with larger size, the class distribution will be close to uniform and each class has roughly equal number of positives. Thus, selecting top-$k$ is more suitable for NCL, which potentially leverages class-balance property and helps a robust training. In our experiment, we find that using top-$k$ for BCE and threshold for NCL lead to lower results.

\begin{table}[!t]
  \centering
  \footnotesize
  \newcolumntype{C}{>{\centering\arraybackslash}X}%
  \newcolumntype{R}{>{\raggedleft\arraybackslash}X}%
  \begin{tabularx}{0.95\linewidth}{l|c|c}
    \hline
    Method   &   CIFAR-10   & CIFAR-100 \\
    \hline
    Baseline & 87.9$\pm$0.7\%& 69.4$\pm$1.4\%\\
    \hline
    NCL & 93.4$\pm$0.2\%~\textcolor{dark-green}{($\uparrow$ 5.5\%)}& 82.3$\pm$2.6\%~\textcolor{dark-green}{($\uparrow$ 12.9\%)}\\
    NCL w/o PP & 61.8$\pm$7.6\%~\textcolor{dark-red}{($\downarrow$ 26.1\%)}& 68.5$\pm$1.9\%~\textcolor{dark-red}{($\downarrow$ 0.9\%)}\\
    NCL + HNG & 93.4$\pm$0.1\%~\textcolor{dark-green}{($\uparrow$ 5.5\%)}& 86.6$\pm$0.4\%~\textcolor{dark-green}{($\uparrow$ 17.2\%)}\\
    \hline
  \end{tabularx}
  \caption{Evaluation of the effectiveness of the proposed neighborhood contrastive learning (NCL) and hard negative generation (HNG). \textbf{NCL w/o PP}: NCL without pseudo-positives.}
  \label{tabel:dis}
\end{table}
%------------------------------------------------------------------------

{\small
\bibliographystyle{ieee_fullname}
\bibliography{egbib}
}

\end{document}